\documentclass{IOS-Book-Article}
\usepackage{todonotes}
\usepackage{graphicx}

\usepackage{mathptmx}
\usepackage{soul}\setuldepth{article}
\def\hb{\hbox to 11.5 cm{}}
\usepackage[
type={CC},
modifier={by},
version={4.0},
]{doclicense}

\begin{document}

\pagestyle{headings}
\def\thepage{}
\begin{frontmatter}              
\newcommand\blfootnote[1]{%
  \begingroup
  \renewcommand\thefootnote{}\footnote{#1}%
  \addtocounter{footnote}{-1}%
  \endgroup
}
\blfootnote{\doclicenseThis}
\title{Justifiable Artificial Intelligence: Engineering Large Language Models for Legal Applications}

\markboth{}{November 2023\hb}

\author[A,B]{\fnms{Sabine} \snm{Wehnert}\orcid{0000-0002-5290-0321}%
\thanks{Corresponding Author: Sabine Wehnert, sabine.wehnert@gei.de}},

\runningauthor{Wehnert}
\address[A]{Leibniz Institute for Educational Media | GEI, Germany}
\address[B]{Otto von Guericke University Magdeburg, Germany}

\begin{abstract}
In this work, I discuss how Large Language Models can be applied in the legal domain, circumventing their current drawbacks. Despite their large success and acceptance, their lack of explainability hinders legal experts to trust in their output, and this happens rightfully so. However, in this paper, I argue in favor of a new view, Justifiable Artificial Intelligence, instead of focusing on Explainable Artificial Intelligence. I discuss in this paper how gaining evidence for and against a Large Language Model's output may make their generated texts more trustworthy - or hold them accountable for misinformation.
\end{abstract}

\end{frontmatter}
\markboth{November 2023\hb}{November 2023\hb}

\section{Introduction}
The release of Large Language Models (LLMs) has changed our reality. Not only researchers use and test the models' capabilities and susceptibilities, but also companies and individuals employ the generated texts for their everyday needs. Despite much criticism regarding the trustworthiness of the models' responses, the popularity of Artificial Intelligence (AI) -based assistants is ever-increasing. However, it has not been disclosed for many of these models which data they have been trained on, making it hard to estimate model bias that could have been learned from the data. Furthermore, the models' responses may consist of passages from copyrighted training data \cite{karamolegkou2023copyright,sag2023copyright}, despite the generative nature of the underlying transformer architecture \cite{DBLP:conf/nips/VaswaniSPUJGKP17}. Although ongoing research targets this issue \cite{chu2023protect}, there are further problems that have not been resolved (yet). For experts in the legal domain, a notable disadvantage of the current deep learning-based systems used in the LLMs is their lack of \emph{explainability}. In many other domains (such as medicine), there is an unnegotiable need for Explainable Artificial Intelligence (XAI). However, the efforts in this very active research direction have not reached the current LLM applications so far. In this paper, I advocate for using LLMs in the legal domain: Instead of insisting on AI to be explainable, we can require the output of the AI to be \emph{justifiable}. Justifiable Artificial Intelligence, as I will explain in detail, relies on providing evidence from trustworthy sources for and against a claim made by the AI. This leaves the user of such a system with an informed choice to either accept or reject the claims made by an AI. 

The contributions of this position paper are the following:
\begin{itemize}
    \item I coin the term \emph{Justifiable Artificial Intelligence}, describing an alternative approach to the ones taken in Explainable Artificial Intelligence.
    \item I connect Justifiable AI to existing state-of-the art research in Large Language models and fact-checking, uncover challenges and their meaning for applications in the legal domain.
\end{itemize}
The remainder of this work is structured as follows:
In Section \ref{sec:background}, I describe the capabilities of Large Language Models, their evolution, and applications in the legal domain using LLMs. Furthermore, I briefly introduce related works in fact-checking.
In Section \ref{sec:JAI}, I explain how \emph{Justifiable Artificial Intelligence} can be constructed.
In Section \ref{sec:related_work}, I collect other uses of the term ``Justifiable Artificial Intelligence".
In Section \ref{sec:conclusion}, I conclude the work and point to possible future work.
\section{Background}\label{sec:background}
\subsection{Large Language Models}\label{subsec:llm}
The aim of language models is to learn probabilities of word sequences, based on different tasks, such as predicting the next sentence or a masked word inside a sequence. Through the exposure to many examples, these models form a representation of the language(s) they are trained on, enabling them to model semantic similarity between words and word sequences.
Large Language Models (LLMs) are large artificial neural networks that have been trained on large collections of data and have billions of parameters. The first notable attempt towards an LLM may be GPT-2, which at the time of its invention was capable of reaching state-of-the-art performance in many natural language generation tasks \cite{radford2019better}. This was considered a breakthrough and because of many concerns regarding its misuse, this model was not released for public use by OpenAI, the company that trained the model. However, this did not hinder further research and development in that direction. Newer versions of GPT are prominent examples of LLMs, which are nowadays accessible to the public as \emph{ChatGPT}\footnote{\url{https://openai.com/blog/chatgpt/}}, with GPT-3.5 or GPT-4 (a multi-modal model, capable of processing images) \cite{DBLP:journals/corr/abs-2303-08774} as the backend. Aside from OpenAI, other companies also released their own versions of LLMs: LaMDA \cite{DBLP:journals/corr/abs-2201-08239}, LLaMa \cite{llama}, MT-NLG~\cite{mtnlg}, PaLM2 \cite{palm2}, OPT \cite{opt}, Sparrow \cite{DBLP:journals/corr/abs-2209-14375}, and many others. An important skill that these models have is in-context learning \cite{DBLP:conf/nips/BrownMRSKDNSSAA20}, meaning that the models can generate text based on a prompt, the so-called context, enabling them to hold interactive conversations with their users. Furthermore, LLMs perform Reinforcement Learning from Human Feedback (RLHF), which is the process of fine-tuning the model based on rewards inferred from their users' responses to their output \cite{DBLP:journals/corr/abs-1909-08593,DBLP:conf/nips/ChristianoLBMLA17,DBLP:conf/nips/Ouyang0JAWMZASR22}. Although LLMs excel in many different tasks without the need for specific training (e.g., Medical Licensing Exams \cite{DBLP:journals/corr/abs-2307-00112,DBLP:journals/corr/abs-2303-08774,thirunavukarasu2023trialling,DBLP:journals/corr/abs-2303-13375}), their training procedure does not equip them with true Natural Language Understanding capabilities comparable to human problem solving. Often, LLMs are attributed with the skill of ``understanding" human language. I do not support this view, since the evidence at the time of writing this paper is not sufficient and in many cases teaching us the opposite (see the debunking \cite{DBLP:journals/corr/abs-2304-15004} of seemingly emergent capabilities coming from model size scaling \cite{DBLP:journals/tmlr/WeiTBRZBYBZMCHVLDF22}). These models are well-versed in modeling language based on many examples of word co-occurrences and plausible text sequences they have been exposed to during the training phase. There are key limitations that many of these language models have \cite{thirunavukarasu2023large}: 
\begin{itemize}
    \item \emph{Recency: } the training data is limited until a specific cutoff day, 
    \item \emph{Accuracy:} the training data stems from online content (websites, books) that has not been thoroughly fact-checked, 
    \item \emph{Coherence: } the generated text is written in a coherent manner, but can be entirely fictional (referred to as ``hallucinations"), when prompted for references, they may be invented \cite{day2023preliminary,giray2023chatgpt},
    \item \emph{Transparency \& Interpretability: } deep neural networks are mostly seen as ``black boxes" by users, visualizing their weights or attention may not be helpful,
    \item \emph{Ethical Concerns: } Privacy issues may arise because the use of personal data may not be revocable, the ``right to be forgotten" cannot be easily enforced when models take months to train. Also, bias and unfairness may be present due to a bias in the training data (e.g., religion bias \cite{DBLP:conf/aies/AbidF021}). Furthermore, the high energy consumption during training and provision of these models needs to be in relation to efficiency gains and potential benefits for the environment \cite{doi:10.1021/acs.est.3c01106}.
\end{itemize}
There are efforts to address some of these issues, such as giving access to the world wide web to BlenderBot 3 \cite{DBLP:journals/corr/abs-2208-03188}, Bing AI \cite{BingAI} and Sparrow \cite{DBLP:journals/corr/abs-2209-14375}. ChatGPT can receive real-time information through APIs\footnote{\url{https://platform.openai.com/docs/plugins/introduction}}. In the future, we may be able to align AI with societal values~\cite{DBLP:journals/corr/abs-2209-13020} and through human feedback we may be able to adjust the biases of an LLM~\cite{DBLP:journals/corr/abs-2302-07459}, solving the ethical concerns in LLMs. Some solutions are proposed for ensuring privacy in LLMs~\cite{DBLP:journals/corr/abs-2307-03941}. 
\subsection{Explainability in Large Language Models}\label{subsec:XAI}
However, the key limitation of Transparency \& Interpretability has not been sufficiently addressed by Explainable AI (XAI) research so far \cite{thirunavukarasu2023large,DBLP:journals/corr/abs-2304-00612}. Compared to other Recurrent Neural Network (RNN) architectures, the transformer architecture \cite{DBLP:conf/nips/VaswaniSPUJGKP17} (i.e., the basis of many LLMs) is relatively easier to interpret through the use of neural attention, however this is often not sufficient as the sole base for transparency, especially in more complex, non-classification tasks, such as translation or natural language inference \cite{10.1145/3539618.3591879}. Possible approaches for explaining transformers are saliency maps, feature centrality scores, and counterfactual explanations \cite{hassija2023interpreting,DBLP:journals/ijhci/SilvaSHGG23}. A more recent way to improve the interpretability of language models is to extract knowledge graphs from them and to compare these representations among different language models \cite{DBLP:journals/corr/abs-2307-00364}, or to generate natural language explanations based on these graphs for a label and against the other alternative(s) \cite{DBLP:journals/corr/abs-2303-16537}. 

As investment in LLMs grows, their capabilities in general may also improve \cite{DBLP:journals/corr/abs-2304-00612}. But how about tasks in the legal domain, which appear to be often rule-based, and - potentially - learnable by an LLM?

\subsection{Large Language Models in the Legal Domain}\label{subsec:lawllm}
A comprehensive survey of the transformer-based language model use in the legal domain has been published by Greco and Tagarelli \cite{DBLP:journals/corr/abs-2308-05502}. In the scope of this work, I only collect the most recent works connected to LLMs to provide a general understanding of the caveats coming with applying LLMs on legal language. The inclined reader may refer to the aforementioned work for a more detailed description. 

As one of the basic legal skills, reasoning has been tested on GPT-3. One type of legal reasoning involves having a pair of facts and a statute, and deciding if the statute applies to those facts. This is often framed as an entailment classification task. Blair-Stanek et al. \cite{DBLP:conf/icail/Blair-StanekHD23} have achieved state-of-the-art performance on such an entailment task, using GPT-3. However, several mis-classifications have led to the conclusion that the task cannot be automated with that LLM because of its incorrect knowledge of the relevant statutes. With guaranteed unseen, synthetic examples, the model also fails to answer the questions correctly. Legal reasoning is a complex task involving many problem-solving capabilities, which go beyond semantic similarity of text. Despite the observation of potential analogical reasoning capabilities in LLMs \cite{DBLP:journals/corr/abs-2212-09196}, there is an ongoing debate and the demand to prove that models do not simply memorize training data instead~\cite{DBLP:journals/corr/abs-2308-16118}. To understand the performance of LLMs on legal reasoning, the benchmark dataset LEGALBENCH has been proposed with 162 tasks, divided in six reasoning categories \cite{DBLP:journals/corr/abs-2308-11462}.

Prompting is an important task to consider when interacting with an LLM. For this, a study \cite{DBLP:journals/corr/abs-2212-01326} has been performed to prompt GPT-3 using the Chain-of-Thought (CoT) technique \cite{DBLP:conf/nips/KojimaGRMI22} on the statute entailment task of the Competition on Legal Information Extraction/Entailment (COLIEE). Despite achieving state-of-the-art performance, the authors express doubt that LLMs can be taught to follow a logical line of thought of a lawyer. Another work about negating prompts revealed weaknesses of GPT-3 to follow the modified instructions properly \cite{DBLP:conf/tl4nlp/JangYS22}. Syllogism prompting (consisting of the law as a major premise, the fact is the minor premise and the judgment as the conclusion) can boost performance in legal judgment prediction \cite{10.1145/3594536.3595170}. Other prompting techniques, such as IRAC (Issue, Rule, Application, Conclusion) have been tested on the COLIEE 2021 and COLIEE 2022 statutory entailment task by Yu et al. \cite{DBLP:conf/acl/YuQS23}. Similarly, Chain of Reference prompting has been successfully applied on COLIEE 2021 statute law entailment~\cite{kuppa2023chain}. For legal case retrieval on the COLIEE 2023 dataset, a framework called PromptCase~\cite{DBLP:journals/corr/abs-2309-02962} has been tested. It consists of issue and fact extraction as the first step, followed by dual and cross encoding with the prompt of a case. 

The T5 model has been employed in COLIEE's legal case entailment task and reached state-of-the-art performance \cite{DBLP:journals/corr/abs-2205-15172,debbarma2023iitdli}. The authors \cite{DBLP:journals/corr/abs-2205-15172} mention the inference speed for legal cases as a drawback of their method. In the recent COLIEE 2023 edition, LLMs have been trained on an augmented dataset of legal cases by the JNLP team \cite{DBLP:conf/icail/GoebelKK0SY23}. There appears to be a general trend of combining traditional methods for information retrieval with LLMs \cite{DBLP:conf/icail/GoebelKK0SY23}.

LLMs have been employed on U.S. court opinions on fiduciary duties and reached a satisfying, but not sufficient performance \cite{DBLP:journals/corr/abs-2301-10095}. Furthermore, LLMs have been used for Legal Judgment Prediction, and the results were below the state of the art, despite efforts to engineer prompts \cite{DBLP:journals/corr/abs-2212-02199}.
Deroy et al. \cite{DBLP:conf/icail/DeroyG023} employ LLMs for summarization of Indian court case judgements and find inconsistencies in the summaries, as well as hallucinations. They conclude that the use of LLMs for their task is only acceptable in a human-in-the-loop setting, given current model capabilities. This finding is in line with a study on ChatGPT (with GPT-3.5), where the model drafted several legal documents, but failed to detect recent legal cases \cite{iu2023chatgpt}.

Further use cases are legal educational settings, where LLMs can be worthwhile assistants, boosting the teachers' creativity and productivity \cite{prof}. There are also considerations regarding law students' use of LLMs in class or in exams, since LLMs have become part of the working culture  \cite{hargreaves2023words}. LLMs have been tested on law exams, and mostly obtained a passing grade \cite{DBLP:journals/corr/abs-2212-14402,choi2023chatgpt}, with the exception of GPT-4, which can be seen to perform in one study on par with average human student performance \cite{katz2023gpt}.

A remarkable approach especially in the context of this work is ChatLaw \cite{DBLP:journals/corr/abs-2306-16092}, a Legal LLM which uses vector knowledge bases together with reference documents, performing keyword search on them using a BERT model to avoid the issue with hallucinations. A similar architecture called DISC-LawLLM \cite{DBLP:journals/corr/abs-2309-11325} has been proposed recently. The use of knowledge bases is a common hallucination mitigation strategy \cite{DBLP:journals/corr/abs-2309-01219}, usually employed to correct the LLM's output without showing the ``raw knowledge" to the user. Louis et al. \cite{DBLP:journals/corr/abs-2309-17050} on the other hand, have proposed an LLM with a retriever module for answering statutory law-related questions, which generates answers using an extractive method to generate the rationale. This is very much in line with what I envision for Justifiable Artificial Intelligence. 

\subsection{Fact-checking}
Fact-checking, when seen as an automated task, consists of a claim that has to be compared against fact-checked or trustworthy sources and either labeled as true or false. Sometimes, there is also a neutral category when not enough information is present to decide the truthfulness of a claim. Nowadays, LLMs are used for many tasks. Fact-checking is no exception. Any research about using LLMs as the only source for fact-checking is out of scope of this work because of the many reasons I listed before on why LLMs may not generate accurate answers (see section \ref{subsec:llm}).

Yao et al. \cite{10.1145/3539618.3591879} publish a multi-modal dataset for fact-checking, using the common websites for gathering their evidence\footnote{\url{https://www.snopes.com}, \url{https://www.politifact.com}}. Having their data manually fact-checked has the advantage of offering a high-quality dataset to train models on. However, their models perform poorly on the test data. In addition, the multi-modal setting is challenging, if applied on a real use case: If taken out of context, an image can be evidence for anything. It requires a careful consideration of the metadata of the image: When and where has the image been taken (in case of a photo), what is the source of the image (in case of a figure displaying data). Indeed, there are many non-trivial legal issues attached to using images as evidence. Therefore, I do not consider the multi-modal use case for our characterization of Justifiable AI, and instead focus on a task based on textual data only. 

An architecture called FacTeR-Check has been proposed \cite{DBLP:journals/kbs/MartinHHVC22} for misinformation tracking or semi-automated fact-checking. In the fact-checking process, FacTeR-Check uses the semantic similarity from transformers for retrieving fact-checked claims from a database, given a claim (e.g., from an online social network). After ranking top-n fact-checked results, they are compared with the claim in question and classified regarding their entailment (positive or negative). Trokhymovych and Saez-Trumper developed the WikiCheck API \cite{DBLP:conf/cikm/TrokhymovychS21} that can be used for fact-checking claims through the Wikipedia knowledge base. Their API uses the MediaWiki API\footnote{\url{https://www.mediawiki.org/wiki/API:Search}} to extract related articles, given a (potentially enriched) claim. Then, entailment is classified using a transformer-based model architecture.

To conclude, fact-checking requires a collection of trusted sources (which is not always easy to determine), a retrieval module and a way to compare the claim and the fact (e.g., via entailment classification). Is fact-checking all we need? This question cannot be answered easily, however following a process based on the general idea of fact-checking may be the answer to the reputation problem LLMs have in the legal domain nowadays.

\section{Justifiable Artificial Intelligence}\label{sec:JAI}

After exploring the state-of-the-art in LLMs, XAI, LLM use in the legal domain, and fact-checking, I come to the conclusion that LLMs are currently not well-equipped to be trusted in by legal experts, and rightfully so. Providing \emph{real} references is a useful step towards enabling a human to check the LLM's output and to trust its validity. Architectures similar to ChatLaw or DISC-LawLLM (previously mentioned in section \ref{subsec:lawllm}) are a good method to increase the accuracy of a model. Depending on the task, it may be worthwhile to offer insight into the references taken by an LLM to the user. This is where approaches with \emph{Justifiable Artificial Intelligence} can fill the gap: Providing evidence to the user, on the base of real documents, shall justify the AI's output. As already mentioned, the work by Louis et al. \cite{DBLP:journals/corr/abs-2309-17050} is an example for this type of approaches. 

However, the justification does not necessarily have to be performed by the LLM itself. Justifiable AI follows the motto: \emph{If you cannot explain yourself, at least justify your opinion.} With that I do not only mean providing evidence backing up a claim, but also (if useful for the task) showing possible evidence against the claim. A retrieval module may search for online content for and against a statement made by the LLM. In that way, the user does not get influenced by a confirmation bias, but can look into different perspectives, when the answer may not be straightforward. This is where the place of LLMs shall be nowadays: Assistants, that inform the user about an issue from different perspectives, but do not influence the user in one direction. We can clearly see that as long as there are issues with \emph{recency, accuracy, coherence, transparency \& interpretability}, as well as \emph{ethical issues}, the use of LLMs in domains that demand all these aspects, will never be fully autonomous and require human validation. Therefore, I advocate for Justifiable AI, for example via in-built fact-checking modules to assist the users who will have to perform the validation anyway. 

\begin{figure}[htbp]
    \centering
   \includegraphics[scale=.2]{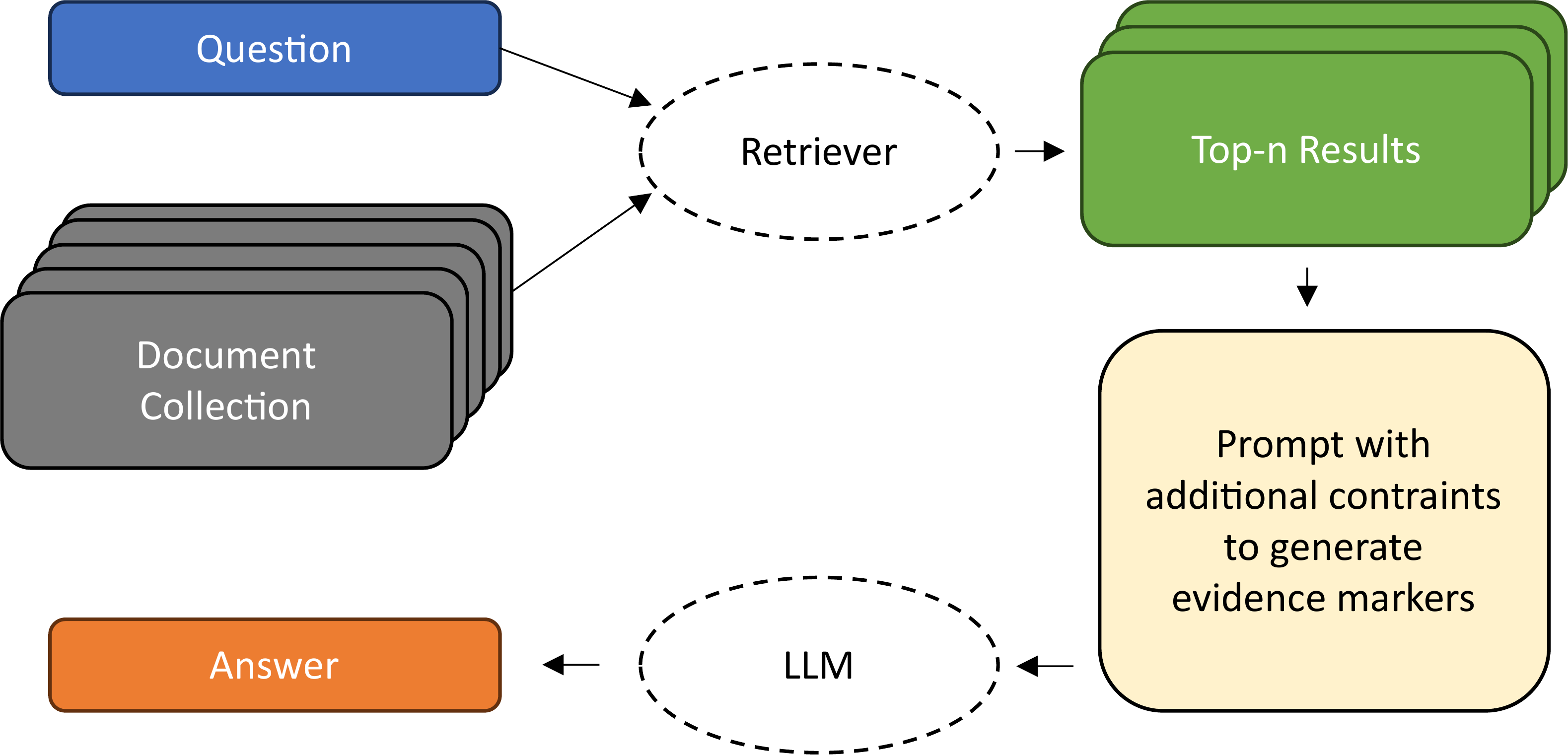}
    \caption{Pipeline for prompting an LLM to extract evidence from documents, adapted from Louis et al. \cite{DBLP:journals/corr/abs-2309-17050}.}
    \label{fig:1}
\end{figure}

\begin{figure}[htbp]
    \centering
   \includegraphics[scale=.2]{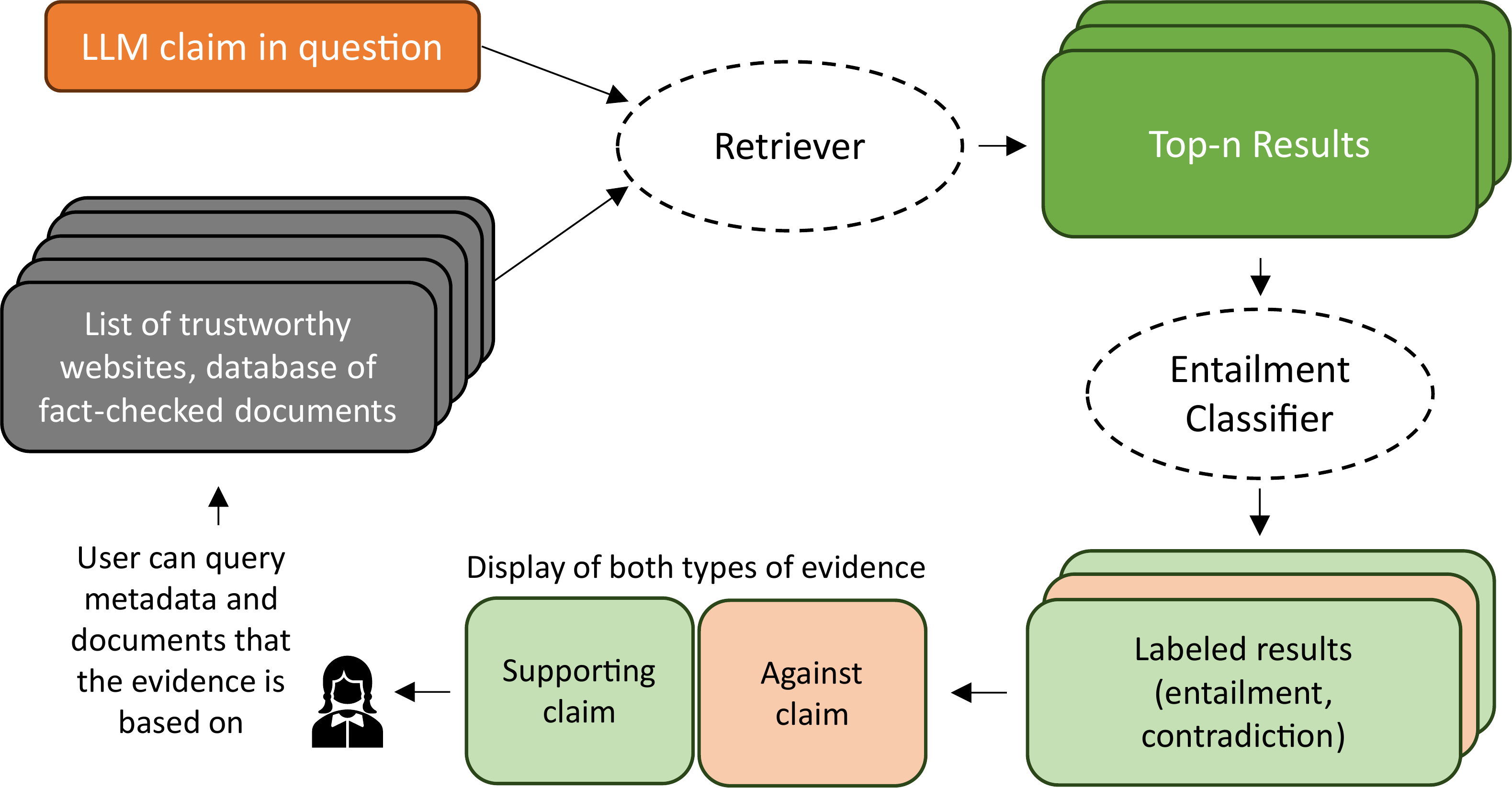}
    \caption{Pipeline for fact-checking a claim made by an LLM} 
    \label{fig:2}
\end{figure}

To illustrate the point, consider the two pipelines of Figure~\ref{fig:1} and Figure~\ref{fig:2}. In Figure~1, we see the workflow by by Louis et al. \cite{DBLP:journals/corr/abs-2309-17050}. This approach can only be accepted as Justifiable AI, if the evidence extraction works properly with an LLM, since this solution is prompt-based. Figure~2 shows the fact-checking based solution, where either trustworthy online sources, or a knowledge base are queried. Then, supporting and contradicting evidence (based on entailment classification) is shown to the user for full decision sovereignty. To ensure transparency, the user can query the document collection (e.g., a database) and understand its composition through metadata and inspection of individual documents.

Note that both approaches are not equal to a fully manual fact-checking process, since prompting, retrieval, and entailment classification can lead to errors that propagate into the final decision making. However, any ``manual" use of a search engine may also not guarantee finding all relevant documents. Therefore, let Justifiable AI be the family of approaches that help us trust LLMs more.

\section{Related Work}\label{sec:related_work}
In this section, I shortly review other uses of the term \emph{Justifiable Artificial Intelligence}. A literal search of the concept ``Justifiable Artificial Intelligence" on google scholar does not return any relevant result. When searching for ``Justifiable AI", I find the word ``justifiable" used as an adjective, often in connection to ``interpretable" or ``morally" (e.g., in this work \cite{DBLP:journals/corr/abs-1906-05684}). Justifiable Artificial Intelligence does not seem to be a proper name used in the research community. It is seen as a property, not as a group of methods (yet). Given the discussion on why current approaches to explain transformer-based outputs may not be helpful to obtain full transparency and interpretability (see section \ref{subsec:XAI}) and the great performance of LLMs met with a lot of caution by legal experts (see section \ref{subsec:lawllm}), emerging methods using \emph{evidence} shall be distinguishable from other works. They are truly human-centered, and will likely be more accepted by legal experts because of the possibility to validate outputs.

Extending the search for the term ``Justifiable Artificial Intelligence" to a regular google search, we find several domains using this term. A commentary by Hadfield \cite{gillian} is about Justifiable AI and that it shall be demanded by law, instead of Explainable AI. She makes the distinction between both worlds with respect to their focus: \emph{``While Explainable AI is focused on fact, justifiable AI is focused on judgment"} \cite{gillian}. This view is more connected to judgment in terms of societal values, instead of the possibility to validate AI through evidence (i.e., what I mean by Justifiable AI). A blog post by Jones is closer to that notion: \emph{``We need to be able to explain how we got to the outcome and really importantly to demonstrate how we were in control of that outcome. In the same way as you can’t explain the individual neurons that fire to enable someone to complete a job successfully, you still need to show how you made sure they did."} \cite{Jones}.

\section{Conclusion}\label{sec:conclusion}
I have demonstrated the current challenges in the use of Large Language Models in the legal domain. As an alternative, I propose to build models that have an additional module for providing extractive rationale or automated retrieval of supporting and contradicting evidence for the model's claims. This enables the users to perform their own fact-checking of the Large Language Models' outputs. The potential of this approach is to encourage legal domain experts to use intelligent assistants in their work without sacrificing the reliability and quality of their results. Future work needs to validate the acceptance of Large Language Models through legal experts, given their outputs are justifiable.

\bibliographystyle{vancouver}
\bibliography{bibliography}

\begin{thebibliography}{10}

\bibitem{karamolegkou2023copyright}
Karamolegkou A, Li J, Zhou L, S{\o}gaard A.
\newblock Copyright Violations and Large Language Models.
\newblock arXiv preprint arXiv:231013771. 2023.

\bibitem{sag2023copyright}
Sag M.
\newblock Copyright safety for generative AI.
\newblock Forthcoming in the Houston Law Review. 2023.

\bibitem{DBLP:conf/nips/VaswaniSPUJGKP17}
Vaswani A, Shazeer N, Parmar N, Uszkoreit J, Jones L, Gomez AN, et~al.
\newblock Attention is All you Need.
\newblock In: Guyon I, von Luxburg U, Bengio S, Wallach HM, Fergus R, Vishwanathan SVN, et~al., editors. Advances in Neural Information Processing Systems 30: Annual Conference on Neural Information Processing Systems 2017, December 4-9, 2017, Long Beach, CA, {USA}; 2017. p. 5998-6008.
\newblock Available from: \url{https://proceedings.neurips.cc/paper/2017/hash/3f5ee243547dee91fbd053c1c4a845aa-Abstract.html}.

\bibitem{chu2023protect}
Chu T, Song Z, Yang C.
\newblock How to Protect Copyright Data in Optimization of Large Language Models?
\newblock arXiv preprint arXiv:230812247. 2023.

\bibitem{radford2019better}
Radford A, Wu J, Amodei D, Amodei D, Clark J, Brundage M, et~al.
\newblock Better language models and their implications.
\newblock OpenAI blog. 2019;1(2).

\bibitem{DBLP:journals/corr/abs-2303-08774}
OpenAI.
\newblock {GPT-4} Technical Report.
\newblock CoRR. 2023;abs/2303.08774.
\newblock Available from: \url{https://doi.org/10.48550/arXiv.2303.08774}.

\bibitem{DBLP:journals/corr/abs-2201-08239}
Thoppilan R, Freitas DD, Hall J, Shazeer N, Kulshreshtha A, Cheng H, et~al.
\newblock LaMDA: Language Models for Dialog Applications.
\newblock CoRR. 2022;abs/2201.08239.
\newblock Available from: \url{https://arxiv.org/abs/2201.08239}.

\bibitem{llama}
Touvron H, Lavril T, Izacard G, Martinet X, Lachaux M, Lacroix T, et~al.
\newblock LLaMA: Open and Efficient Foundation Language Models.
\newblock CoRR. 2023;abs/2302.13971.
\newblock Available from: \url{https://doi.org/10.48550/arXiv.2302.13971}.

\bibitem{mtnlg}
Smith S, Patwary M, Norick B, LeGresley P, Rajbhandari S, Casper J, et~al.
\newblock Using DeepSpeed and Megatron to Train Megatron-Turing {NLG} 530B, {A} Large-Scale Generative Language Model.
\newblock CoRR. 2022;abs/2201.11990.
\newblock Available from: \url{https://arxiv.org/abs/2201.11990}.

\bibitem{palm2}
Anil R, Dai AM, Firat O, Johnson M, Lepikhin D, Passos A, et~al.
\newblock PaLM 2 Technical Report.
\newblock CoRR. 2023;abs/2305.10403.
\newblock Available from: \url{https://doi.org/10.48550/arXiv.2305.10403}.

\bibitem{opt}
Zhang S, Roller S, Goyal N, Artetxe M, Chen M, Chen S, et~al.
\newblock {OPT:} Open Pre-trained Transformer Language Models.
\newblock CoRR. 2022;abs/2205.01068.
\newblock Available from: \url{https://doi.org/10.48550/arXiv.2205.01068}.

\bibitem{DBLP:journals/corr/abs-2209-14375}
Glaese A, McAleese N, Trebacz M, Aslanides J, Firoiu V, Ewalds T, et~al.
\newblock Improving alignment of dialogue agents via targeted human judgements.
\newblock CoRR. 2022;abs/2209.14375.
\newblock Available from: \url{https://doi.org/10.48550/arXiv.2209.14375}.

\bibitem{DBLP:conf/nips/BrownMRSKDNSSAA20}
Brown TB, Mann B, Ryder N, Subbiah M, Kaplan J, Dhariwal P, et~al.
\newblock Language Models are Few-Shot Learners.
\newblock In: Larochelle H, Ranzato M, Hadsell R, Balcan M, Lin H, editors. Advances in Neural Information Processing Systems 33: Annual Conference on Neural Information Processing Systems 2020, NeurIPS 2020, December 6-12, 2020, virtual; 2020. Available from: \url{https://proceedings.neurips.cc/paper/2020/hash/1457c0d6bfcb4967418bfb8ac142f64a-Abstract.html}.

\bibitem{DBLP:journals/corr/abs-1909-08593}
Ziegler DM, Stiennon N, Wu J, Brown TB, Radford A, Amodei D, et~al.
\newblock Fine-Tuning Language Models from Human Preferences.
\newblock CoRR. 2019;abs/1909.08593.
\newblock Available from: \url{http://arxiv.org/abs/1909.08593}.

\bibitem{DBLP:conf/nips/ChristianoLBMLA17}
Christiano PF, Leike J, Brown TB, Martic M, Legg S, Amodei D.
\newblock Deep Reinforcement Learning from Human Preferences.
\newblock In: Guyon I, von Luxburg U, Bengio S, Wallach HM, Fergus R, Vishwanathan SVN, et~al., editors. Advances in Neural Information Processing Systems 30: Annual Conference on Neural Information Processing Systems 2017, December 4-9, 2017, Long Beach, CA, {USA}; 2017. p. 4299-307.
\newblock Available from: \url{https://proceedings.neurips.cc/paper/2017/hash/d5e2c0adad503c91f91df240d0cd4e49-Abstract.html}.

\bibitem{DBLP:conf/nips/Ouyang0JAWMZASR22}
Ouyang L, Wu J, Jiang X, Almeida D, Wainwright CL, Mishkin P, et~al.
\newblock Training language models to follow instructions with human feedback.
\newblock In: NeurIPS; 2022. Available from: \url{http://papers.nips.cc/paper\_files/paper/2022/hash/b1efde53be364a73914f58805a001731-Abstract-Conference.html}.

\bibitem{DBLP:journals/corr/abs-2307-00112}
Sharma P, Thapa K, Thapa D, Dhakal P, Upadhaya MD, Adhikari S, et~al.
\newblock Performance of ChatGPT on {USMLE:} Unlocking the Potential of Large Language Models for AI-Assisted Medical Education.
\newblock CoRR. 2023;abs/2307.00112.
\newblock Available from: \url{https://doi.org/10.48550/arXiv.2307.00112}.

\bibitem{thirunavukarasu2023trialling}
Thirunavukarasu AJ, Hassan R, Mahmood S, Sanghera R, Barzangi K, El~Mukashfi M, et~al.
\newblock Trialling a large language model (ChatGPT) in general practice with the Applied Knowledge Test: observational study demonstrating opportunities and limitations in primary care.
\newblock JMIR Medical Education. 2023;9(1):e46599.

\bibitem{DBLP:journals/corr/abs-2303-13375}
Nori H, King N, McKinney SM, Carignan D, Horvitz E.
\newblock Capabilities of {GPT-4} on Medical Challenge Problems.
\newblock CoRR. 2023;abs/2303.13375.
\newblock Available from: \url{https://doi.org/10.48550/arXiv.2303.13375}.

\bibitem{DBLP:journals/corr/abs-2304-15004}
Schaeffer R, Miranda B, Koyejo S.
\newblock Are Emergent Abilities of Large Language Models a Mirage?
\newblock CoRR. 2023;abs/2304.15004.
\newblock Available from: \url{https://doi.org/10.48550/arXiv.2304.15004}.

\bibitem{DBLP:journals/tmlr/WeiTBRZBYBZMCHVLDF22}
Wei J, Tay Y, Bommasani R, Raffel C, Zoph B, Borgeaud S, et~al.
\newblock Emergent Abilities of Large Language Models.
\newblock Trans Mach Learn Res. 2022;2022.
\newblock Available from: \url{https://openreview.net/forum?id=yzkSU5zdwD}.

\bibitem{thirunavukarasu2023large}
Thirunavukarasu AJ, Ting DSJ, Elangovan K, Gutierrez L, Tan TF, Ting DSW.
\newblock Large language models in medicine.
\newblock Nature medicine. 2023;29(8):1930-40.

\bibitem{day2023preliminary}
Day T.
\newblock A preliminary investigation of fake peer-reviewed citations and references generated by ChatGPT.
\newblock The Professional Geographer. 2023:1-4.

\bibitem{giray2023chatgpt}
Giray L.
\newblock ChatGPT References Unveiled: Distinguishing the Reliable from the Fake.
\newblock Internet Reference Services Quarterly. 2023:1-10.

\bibitem{DBLP:conf/aies/AbidF021}
Abid A, Farooqi M, Zou J.
\newblock Persistent Anti-Muslim Bias in Large Language Models.
\newblock In: Fourcade M, Kuipers B, Lazar S, Mulligan DK, editors. {AIES} '21: {AAAI/ACM} Conference on AI, Ethics, and Society, Virtual Event, USA, May 19-21, 2021. {ACM}; 2021. p. 298-306.
\newblock Available from: \url{https://doi.org/10.1145/3461702.3462624}.

\bibitem{doi:10.1021/acs.est.3c01106}
Rillig MC, Ågerstrand M, Bi M, Gould KA, Sauerland U.
\newblock Risks and Benefits of Large Language Models for the Environment.
\newblock Environmental Science \& Technology. 2023;57(9):3464-6.
\newblock PMID: 36821477.
\newblock Available from: \url{https://doi.org/10.1021/acs.est.3c01106}.

\bibitem{DBLP:journals/corr/abs-2208-03188}
Shuster K, Xu J, Komeili M, Ju D, Smith EM, Roller S, et~al.
\newblock BlenderBot 3: a deployed conversational agent that continually learns to responsibly engage.
\newblock CoRR. 2022;abs/2208.03188.
\newblock Available from: \url{https://doi.org/10.48550/arXiv.2208.03188}.

\bibitem{BingAI}
Mehdi Y. Reinventing search with a new AI-powered Microsoft Bing and Edge, your copilot for the web; 2023.
\newblock Accessed on: 02 November 2023.
\newblock Available from: \url{https://tinyurl.com/yckmtuss}.

\bibitem{DBLP:journals/corr/abs-2209-13020}
Nay JJ.
\newblock Law Informs Code: {A} Legal Informatics Approach to Aligning Artificial Intelligence with Humans.
\newblock CoRR. 2022;abs/2209.13020.
\newblock Available from: \url{https://doi.org/10.48550/arXiv.2209.13020}.

\bibitem{DBLP:journals/corr/abs-2302-07459}
Ganguli D, Askell A, Schiefer N, Liao TI, Lukosiute K, Chen A, et~al.
\newblock The Capacity for Moral Self-Correction in Large Language Models.
\newblock CoRR. 2023;abs/2302.07459.
\newblock Available from: \url{https://doi.org/10.48550/arXiv.2302.07459}.

\bibitem{DBLP:journals/corr/abs-2307-03941}
Zhang D, Finckenberg{-}Broman P, Hoang T, Pan S, Xing Z, Staples M, et~al.
\newblock Right to be Forgotten in the Era of Large Language Models: Implications, Challenges, and Solutions.
\newblock CoRR. 2023;abs/2307.03941.
\newblock Available from: \url{https://doi.org/10.48550/arXiv.2307.03941}.

\bibitem{DBLP:journals/corr/abs-2304-00612}
Bowman SR.
\newblock Eight Things to Know about Large Language Models.
\newblock CoRR. 2023;abs/2304.00612.
\newblock Available from: \url{https://doi.org/10.48550/arXiv.2304.00612}.

\bibitem{10.1145/3539618.3591879}
Yao BM, Shah A, Sun L, Cho JH, Huang L.
\newblock End-to-End Multimodal Fact-Checking and Explanation Generation: A Challenging Dataset and Models.
\newblock SIGIR '23. New York, NY, USA: Association for Computing Machinery; 2023. p. 2733–2743.
\newblock Available from: \url{https://doi.org/10.1145/3539618.3591879}.

\bibitem{hassija2023interpreting}
Hassija V, Chamola V, Mahapatra A, Singal A, Goel D, Huang K, et~al.
\newblock Interpreting black-box models: a review on explainable artificial intelligence.
\newblock Cognitive Computation. 2023:1-30.

\bibitem{DBLP:journals/ijhci/SilvaSHGG23}
Silva A, Schrum M, Hedlund{-}Botti E, Gopalan N, Gombolay MC.
\newblock Explainable Artificial Intelligence: Evaluating the Objective and Subjective Impacts of xAI on Human-Agent Interaction.
\newblock Int J Hum Comput Interact. 2023;39(7):1390-404.
\newblock Available from: \url{https://doi.org/10.1080/10447318.2022.2101698}.

\bibitem{DBLP:journals/corr/abs-2307-00364}
Swamy V, Frej J, K{\"{a}}ser T.
\newblock The future of human-centric eXplainable Artificial Intelligence {(XAI)} is not post-hoc explanations.
\newblock CoRR. 2023;abs/2307.00364.
\newblock Available from: \url{https://doi.org/10.48550/arXiv.2307.00364}.

\bibitem{DBLP:journals/corr/abs-2303-16537}
Chen Z, Singh AK, Sra M.
\newblock LMExplainer: a Knowledge-Enhanced Explainer for Language Models.
\newblock CoRR. 2023;abs/2303.16537.
\newblock Available from: \url{https://doi.org/10.48550/arXiv.2303.16537}.

\bibitem{DBLP:journals/corr/abs-2308-05502}
Greco CM, Tagarelli A.
\newblock Bringing order into the realm of Transformer-based language models for artificial intelligence and law.
\newblock CoRR. 2023;abs/2308.05502.
\newblock Available from: \url{https://doi.org/10.48550/arXiv.2308.05502}.

\bibitem{DBLP:conf/icail/Blair-StanekHD23}
Blair{-}Stanek A, Holzenberger N, Durme BV.
\newblock Can {GPT-3} Perform Statutory Reasoning?
\newblock In: Grabmair M, Andrade F, Novais P, editors. Proceedings of the Nineteenth International Conference on Artificial Intelligence and Law, {ICAIL} 2023, Braga, Portugal, June 19-23, 2023. {ACM}; 2023. p. 22-31.
\newblock Available from: \url{https://doi.org/10.1145/3594536.3595163}.

\bibitem{DBLP:journals/corr/abs-2212-09196}
Webb TW, Holyoak KJ, Lu H.
\newblock Emergent Analogical Reasoning in Large Language Models.
\newblock CoRR. 2022;abs/2212.09196.
\newblock Available from: \url{https://doi.org/10.48550/arXiv.2212.09196}.

\bibitem{DBLP:journals/corr/abs-2308-16118}
Hodel D, West J.
\newblock Response: Emergent analogical reasoning in large language models.
\newblock CoRR. 2023;abs/2308.16118.
\newblock Available from: \url{https://doi.org/10.48550/arXiv.2308.16118}.

\bibitem{DBLP:journals/corr/abs-2308-11462}
Guha N, Nyarko J, Ho DE, R{\'{e}} C, Chilton A, Narayana A, et~al.
\newblock LegalBench: {A} Collaboratively Built Benchmark for Measuring Legal Reasoning in Large Language Models.
\newblock CoRR. 2023;abs/2308.11462.
\newblock Available from: \url{https://doi.org/10.48550/arXiv.2308.11462}.

\bibitem{DBLP:journals/corr/abs-2212-01326}
Yu F, Quartey L, Schilder F.
\newblock Legal Prompting: Teaching a Language Model to Think Like a Lawyer.
\newblock CoRR. 2022;abs/2212.01326.
\newblock Available from: \url{https://doi.org/10.48550/arXiv.2212.01326}.

\bibitem{DBLP:conf/nips/KojimaGRMI22}
Kojima T, Gu SS, Reid M, Matsuo Y, Iwasawa Y.
\newblock Large Language Models are Zero-Shot Reasoners.
\newblock In: NeurIPS; 2022. Available from: \url{http://papers.nips.cc/paper\_files/paper/2022/hash/8bb0d291acd4acf06ef112099c16f326-Abstract-Conference.html}.

\bibitem{DBLP:conf/tl4nlp/JangYS22}
Jang J, Ye S, Seo M.
\newblock Can Large Language Models Truly Understand Prompts? {A} Case Study with Negated Prompts.
\newblock In: Albalak A, Zhou C, Raffel C, Ramachandran D, Ruder S, Ma X, editors. Transfer Learning for Natural Language Processing Workshop, 03 December 2022, New Orleans, Louisiana, {USA}. vol. 203 of Proceedings of Machine Learning Research. {PMLR}; 2022. p. 52-62.
\newblock Available from: \url{https://proceedings.mlr.press/v203/jang23a.html}.

\bibitem{10.1145/3594536.3595170}
Jiang C, Yang X.
\newblock Legal Syllogism Prompting: Teaching Large Language Models for Legal Judgment Prediction.
\newblock In: Proceedings of the Nineteenth International Conference on Artificial Intelligence and Law. ICAIL '23. New York, NY, USA: Association for Computing Machinery; 2023. p. 417–421.
\newblock Available from: \url{https://doi.org/10.1145/3594536.3595170}.

\bibitem{DBLP:conf/acl/YuQS23}
Yu F, Quartey L, Schilder F.
\newblock Exploring the Effectiveness of Prompt Engineering for Legal Reasoning Tasks.
\newblock In: Rogers A, Boyd{-}Graber JL, Okazaki N, editors. Findings of the Association for Computational Linguistics: {ACL} 2023, Toronto, Canada, July 9-14, 2023. Association for Computational Linguistics; 2023. p. 13582-96.
\newblock Available from: \url{https://doi.org/10.18653/v1/2023.findings-acl.858}.

\bibitem{kuppa2023chain}
Kuppa A, Rasumov-Rahe N, Voses M.
\newblock Chain Of Reference prompting helps LLM to think like a lawyer.
\newblock In: Generative AI+ Law Workshop; 2023. .

\bibitem{DBLP:journals/corr/abs-2309-02962}
Tang Y, Qiu R, Li X.
\newblock Prompt-based Effective Input Reformulation for Legal Case Retrieval.
\newblock CoRR. 2023;abs/2309.02962.
\newblock Available from: \url{https://doi.org/10.48550/arXiv.2309.02962}.

\bibitem{DBLP:journals/corr/abs-2205-15172}
Rosa GM, Bonifacio LH, Jeronymo V, Abonizio HQ, de~Alencar~Lotufo R, Nogueira RF.
\newblock Billions of Parameters Are Worth More Than In-domain Training Data: {A} case study in the Legal Case Entailment Task.
\newblock CoRR. 2022;abs/2205.15172.
\newblock Available from: \url{https://doi.org/10.48550/arXiv.2205.15172}.

\bibitem{debbarma2023iitdli}
Debbarma R, Prawar P, Chakraborty A, Bedathur S.
\newblock IITDLI: Legal Case Retrieval Based on Lexical Models.
\newblock In: Workshop of the Tenth Competition on Legal Information Extraction/Entailment (COLIEE'2023) in the 19th International Conference on Artificial Intelligence and Law (ICAIL); 2023. .

\bibitem{DBLP:conf/icail/GoebelKK0SY23}
Goebel R, Kano Y, Kim M, Rabelo J, Satoh K, Yoshioka M.
\newblock Summary of the Competition on Legal Information, Extraction/Entailment {(COLIEE)} 2023.
\newblock In: Grabmair M, Andrade F, Novais P, editors. Proceedings of the Nineteenth International Conference on Artificial Intelligence and Law, {ICAIL} 2023, Braga, Portugal, June 19-23, 2023. {ACM}; 2023. p. 472-80.
\newblock Available from: \url{https://doi.org/10.1145/3594536.3595176}.

\bibitem{DBLP:journals/corr/abs-2301-10095}
Nay JJ.
\newblock Large Language Models as Fiduciaries: {A} Case Study Toward Robustly Communicating With Artificial Intelligence Through Legal Standards.
\newblock CoRR. 2023;abs/2301.10095.
\newblock Available from: \url{https://doi.org/10.48550/arXiv.2301.10095}.

\bibitem{DBLP:journals/corr/abs-2212-02199}
Trautmann D, Petrova A, Schilder F.
\newblock Legal Prompt Engineering for Multilingual Legal Judgement Prediction.
\newblock CoRR. 2022;abs/2212.02199.
\newblock Available from: \url{https://doi.org/10.48550/arXiv.2212.02199}.

\bibitem{DBLP:conf/icail/DeroyG023}
Deroy A, Ghosh K, Ghosh S.
\newblock How Ready are Pre-trained Abstractive Models and LLMs for Legal Case Judgement Summarization?
\newblock In: Conrad JG, Jr DWL, Baron JR, Henseler H, Bhattacharya P, Nielsen A, et~al., editors. Proceedings of the Third International Workshop on Artificial Intelligence and Intelligent Assistance for Legal Professionals in the Digital Workplace (LegalAIIA 2023) co-located with the 19th International Conference on Artificial Intelligence and Law {(ICAIL} 2023), Braga, Portugal, June 19, 2023. vol. 3423 of {CEUR} Workshop Proceedings. CEUR-WS.org; 2023. p. 8-19.
\newblock Available from: \url{https://ceur-ws.org/Vol-3423/paper2.pdf}.

\bibitem{iu2023chatgpt}
Iu KY, Wong VMY.
\newblock ChatGPT by OpenAI: The End of Litigation Lawyers?
\newblock Available at SSRN. 2023.

\bibitem{prof}
Pettinato~Oltz T.
\newblock ChatGPT, Professor of Law.
\newblock Available at SSRN. 2023.
\newblock Available from: \url{https://ssrn.com/abstract=4347630}.

\bibitem{hargreaves2023words}
Hargreaves S.
\newblock ‘Words Are Flowing Out Like Endless Rain Into a Paper Cup’: ChatGPT \& Law School Assessments.
\newblock The Chinese University of Hong Kong Faculty of Law Research Paper. 2023;(2023-03).

\bibitem{DBLP:journals/corr/abs-2212-14402}
II MJB, Katz DM.
\newblock {GPT} Takes the Bar Exam.
\newblock CoRR. 2022;abs/2212.14402.
\newblock Available from: \url{https://doi.org/10.48550/arXiv.2212.14402}.

\bibitem{choi2023chatgpt}
Choi JH, Hickman KE, Monahan A, Schwarcz D.
\newblock Chatgpt goes to law school.
\newblock Available at SSRN. 2023.

\bibitem{katz2023gpt}
Katz DM, Bommarito MJ, Gao S, Arredondo P.
\newblock Gpt-4 passes the bar exam.
\newblock Available at SSRN 4389233. 2023.

\bibitem{DBLP:journals/corr/abs-2306-16092}
Cui J, Li Z, Yan Y, Chen B, Yuan L.
\newblock ChatLaw: Open-Source Legal Large Language Model with Integrated External Knowledge Bases.
\newblock CoRR. 2023;abs/2306.16092.
\newblock Available from: \url{https://doi.org/10.48550/arXiv.2306.16092}.

\bibitem{DBLP:journals/corr/abs-2309-11325}
Yue S, Chen W, Wang S, Li B, Shen C, Liu S, et~al.
\newblock DISC-LawLLM: Fine-tuning Large Language Models for Intelligent Legal Services.
\newblock CoRR. 2023;abs/2309.11325.
\newblock Available from: \url{https://doi.org/10.48550/arXiv.2309.11325}.

\bibitem{DBLP:journals/corr/abs-2309-01219}
Zhang Y, Li Y, Cui L, Cai D, Liu L, Fu T, et~al.
\newblock Siren's Song in the {AI} Ocean: {A} Survey on Hallucination in Large Language Models.
\newblock CoRR. 2023;abs/2309.01219.
\newblock Available from: \url{https://doi.org/10.48550/arXiv.2309.01219}.

\bibitem{DBLP:journals/corr/abs-2309-17050}
Louis A, van Dijck G, Spanakis G.
\newblock Interpretable Long-Form Legal Question Answering with Retrieval-Augmented Large Language Models.
\newblock CoRR. 2023;abs/2309.17050.
\newblock Available from: \url{https://doi.org/10.48550/arXiv.2309.17050}.

\bibitem{DBLP:journals/kbs/MartinHHVC22}
Mart{\'{\i}}n A, Huertas{-}Tato J, Huertas{-}Garc{\'{\i}}a {\'{A}}, Villar{-}Rodr{\'{\i}}guez G, Camacho D.
\newblock FacTeR-Check: Semi-automated fact-checking through semantic similarity and natural language inference.
\newblock Knowl Based Syst. 2022;251:109265.
\newblock Available from: \url{https://doi.org/10.1016/j.knosys.2022.109265}.

\bibitem{DBLP:conf/cikm/TrokhymovychS21}
Trokhymovych M, S{\'{a}}ez{-}Trumper D.
\newblock WikiCheck: An End-to-end Open Source Automatic Fact-Checking {API} based on Wikipedia.
\newblock In: Demartini G, Zuccon G, Culpepper JS, Huang Z, Tong H, editors. {CIKM} '21: The 30th {ACM} International Conference on Information and Knowledge Management, Virtual Event, Queensland, Australia, November 1 - 5, 2021. {ACM}; 2021. p. 4155-64.
\newblock Available from: \url{https://doi.org/10.1145/3459637.3481961}.

\bibitem{DBLP:journals/corr/abs-1906-05684}
Leslie D.
\newblock Understanding artificial intelligence ethics and safety.
\newblock CoRR. 2019;abs/1906.05684.
\newblock Available from: \url{http://arxiv.org/abs/1906.05684}.

\bibitem{gillian}
Hadfield GK. Explanation and justification: AI decision-making, law, and the rights of citizens; 2021.
\newblock Available from: \url{https://srinstitute.utoronto.ca/news/hadfield-justifiable-ai}.

\bibitem{Jones}
Jones S. Explainable AI is dead, long live justifiable decisions. Medium; 2023.
\newblock Available from: \url{https://tinyurl.com/mrytfuzs}.

\end{thebibliography}
\end{document}